\documentclass[10pt,twocolumn,letterpaper]{article}

\usepackage[pagenumbers]{cvpr} 

\usepackage{graphicx}
\usepackage{amsmath}
\usepackage{amssymb}
\usepackage{booktabs}

\usepackage{algorithm}
\usepackage{algorithmic}
\usepackage{multirow}
\usepackage{color}

\usepackage[pagebackref,breaklinks,colorlinks]{hyperref}

\usepackage[capitalize]{cleveref}
\crefname{section}{Sec.}{Secs.}
\Crefname{section}{Section}{Sections}
\Crefname{table}{Table}{Tables}
\crefname{table}{Tab.}{Tabs.}

\def\cvprPaperID{7512} 
\def\confName{CVPR}
\def\confYear{2022}

\begin{document}

\title{MetaDT: Meta Decision Tree for Interpretable Few-Shot Learning}

\author{Baoquan Zhang, Hao Jiang, Xutao Li, Shanshan Feng, Yunming Ye\thanks{Corresponding author.}, Rui Ye\\
Harbin Institute of Technology, Shenzhen\\
{\tt\small \{zhangbaoquan, haojiang\}@stu.hit.edu.cn,
	\{lixutao, victor\_fengss, yeyunming\}@hit.edu.cn,} \\
{\tt\small yerui\_hitsz@163.com}
}
\maketitle

\begin{abstract}
   Few-Shot Learning (FSL) is a challenging task, which aims to recognize novel classes with few examples. Recently, lots of methods have been proposed from the perspective of meta-learning and representation learning for improving FSL performance. However, few works focus on the interpretability of FSL decision process. In this paper, we take a step towards the interpretable FSL by proposing a novel decision tree-based meta-learning framework, namely, MetaDT. Our insight is replacing the last black-box FSL classifier of the existing representation learning methods by an interpretable decision tree with meta-learning. The key challenge is how to effectively learn the decision tree (i.e., the tree structure and the parameters of each node) in the FSL setting. To address the challenge, we introduce a tree-like class hierarchy as our prior: 1) the hierarchy is directly employed as the tree structure; 2) by regarding the class hierarchy as an undirected graph, a graph convolution-based decision tree inference network is designed as our meta-learner to learn to infer the parameters of each node. At last, a two-loop optimization mechanism is incorporated into our framework for a fast adaptation of the decision tree with few examples. Extensive experiments on performance comparison and interpretability analysis show the effectiveness and superiority of our MetaDT. Our code will be publicly available upon acceptance. 
   
\end{abstract}

\section{Introduction}
\label{sec_1}

Deep convolutional neural network (CNN) has achieved great success on image classification with abundant labeled data \cite{he2016deep}. However, for many rare or new-found objects, acquiring so much labeled data is unrealistic, which limits their applications in practical scenarios such as drug discovery \cite{altae2017low} and cold-start recommendations \cite{VartakTMBL17}. In contrast, humans can quickly learn and recognize novel classes from very few observations. To bridge the gap, Few-Shot Learning (FSL) problem has been proposed and has attracted wide attention recently. It targets at learning transferable knowledge from some base classes with sufficient labeled samples, and then transferring the knowledge to quickly learn a classifier for novel classes with few examples \cite{WangYKN20}.

\begin{figure}
	\centering
	\begin{subfigure}{1.0\linewidth}
		\includegraphics[width=1.0\columnwidth]{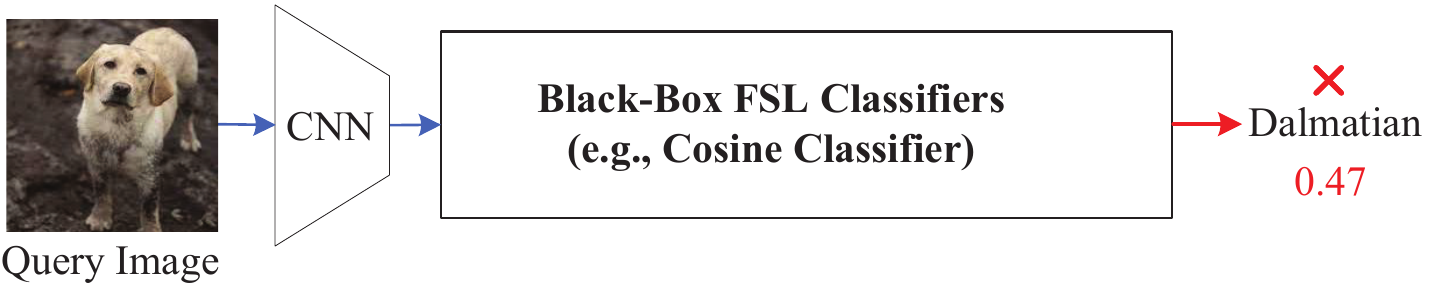}
		\caption{Existing representation learning-based methods.}
		\label{fig1a}
	\end{subfigure}
	\hfill
	\begin{subfigure}{1.0\linewidth}
		\includegraphics[width=1.0\columnwidth]{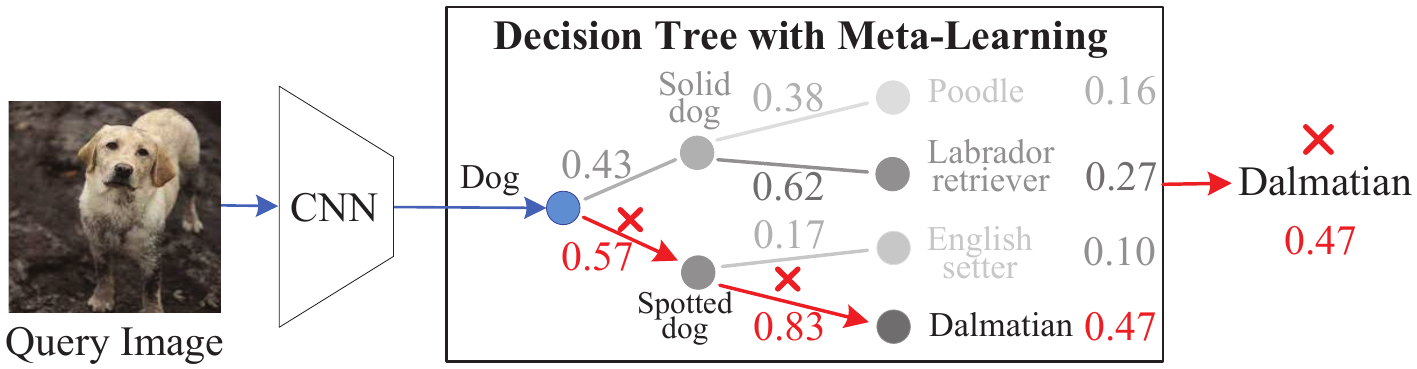}
		\caption{Our MetaDT.}
		\label{fig1b}
	\end{subfigure}
	\caption{Different from existing representation learning methods (a), our MetaDT replaces its black-box classifier by a decision tree with meta-learning (b), which provides interpretability for FSL.
	}
	\vspace{-15pt}
	\label{fig1}
\end{figure}
 
To address the FSL problem, various methods have been proposed, which can be roughly divided into two categories: meta-learning based methods and representation learning-based methods. The former aims to learn a task-agnostic meta-learner (\emph{e.g.}, a good initial model \cite{finn2017model}) by constructing a large number of few-shot tasks from base classes, and then leverages the meta-learner to quickly learn/infer a classifier for novel classes. The latter aims to learn transferable representations from base classes by designing a good training strategy \cite{Rizve_2021_CVPR} or feature extractor \cite{0001FLWLH0X21}, so that the novel classes can be nicely recognized via a simple classifier (\emph{e.g.}, cosine classifier \cite{chen2020new}). Though both types of methods have achieved promising performance on FSL, they mainly focus on improving FSL performance, but have not paid sufficient attention to the interpretability of FSL. In fact, such ability is very important in risk-sensitive applications (\emph{e.g.}, medical diagnosis \cite{banegas2021towards}), where the model not only needs to make a decision (\emph{e.g.}, the clinical outcome of a patient), but also to explain the reason of making such decision.

Recently, several works \cite{CaoBL21, xue2020region, WangLVNKN21} attempt to explore interpretable FSL, \emph{i.e.}, recognizing novel classes meanwhile providing a clear decision explanation for making the class prediction. However, almost all methods focus on explaining how FSL models work from the perspective of heatmap visualization, \emph{e.g.}, visualizing the importance of image regions \cite{xue2020region}. These methods only roughly locate and visualize the image region that the FSL model focuses on, but it is still unclear why the FSL model makes such decision. 

To address the drawback, in this paper, we present a new perspective for interpretable FSL by proposing a novel decision tree-based meta-learning framework, namely, MetaDT. As shown in Figure~\ref{fig1a} and \ref{fig1b}, our key idea is {\bf 1)} replacing the last black-box FSL classifier (\emph{e.g.}, cosine classifier) of the representation learning methods by an interpretable decision tree with meta-learning;
{\bf 2)} introducing a tree-like class hierarchy consisting of few-shot classes (\emph{e.g.}, ``English setter'' or ``Dalmatian'') and their superclasses (\emph{e.g.}, ``Spotted dog'') as the decision tree, where each node in the tree is a learnable model making each step decision 
(\emph{e.g.}, ``Is spotted dog?''); and {\bf 3)} performing novel class prediction in a sequential decision manner from the root of decision tree to its leafs. The advantage of such design is that a sequence of intermediate decisions that lead up to a final class prediction can be obtained, which helps to diagnose why the FSL model makes such prediction. For example, as shown in Figure.~\ref{fig1b}, the image of ``Labrador retriever'' is wrongly predicted to ``Dalmatian''. By examining the intermediate decisions, we can identify the reason for such wrong decision is that the image is misclassified as ``Spotted dog'' when making the decision of ``Is spotted dog or solid dog?''. 

Specifically, the key challenge of such idea is how to effectively learn the decision tree (\emph{i.e.}, the tree structure and the parameters of each node) in the FSL setting. To address the challenge, we regard the introduced class hierarchy as a determined structure prior of the decision tree. By viewing it as an undirected graph, we design a graph convolution-based decision tree inference network as our meta-learner to learn to infer the parameters of each node. The advantage of such design is the class hierarchy relations and abundant class semantics can be fully and effectively exploited in our meta-learner.
Besides, we incorporate a two-loop optimization mechanism \cite{finn2017model} into our framework for a fast adaptation. The outer-loop optimization aims to learn good initial parameters for our meta-learner from abundant base classes. Then, the inner-loop optimization applies the initial parameters to novel classes for quickly inferring a task-specific decision tree with few examples. At last, we perform the novel class prediction in a sequential decision manner from the root of the task-specific decision tree to its leaf nodes.  
 
Our main contributions can be summarized as follows:

\begin{itemize}
	\item We present a new perspective for interpretable FSL by replacing the black-box FSL classifier with an interpretable decision tree, which provides clear decision explanations for class prediction. To our knowledge, this is the first work to explore decision trees on FSL.
	\item We propose a novel decision tree-based meta-learning framework. Here, a graph convolution-based decision tree inference network with class hierarchy is carefully designed for effectively learning a decision tree. Its advantage is the priors of class hierarchy (\emph{i.e.}, class semantics and hierarchy relations) can be fully exploited. 
	\item We conduct comprehensive experiments on miniImagenet, CIFAR-FS, and tieredImagenet, which verify the effectiveness of our MetaDT. In addition, extensive interpretability and visualization analyses show the superiority of our MetaDT for interpretable FSL.
\end{itemize}

\section{Related Work}
\label{sec_2}

\subsection{Few-Shot Learning}
Few-Shot Learning (FSL) aims to recognize novel classes with few labeled samples. Existing methods can be roughly grouped into two categories: meta-learning based approaches and representation learning-based approaches. The meta-learning based approaches focus on learning a task-agnostic meta-learner by constructing a large number of few-shot tasks from base classes, and then leverage the meta-learner to quickly learn/infer an FSL classifier for recognizing novel classes. The meta-learner can be a good initial model \cite{finn2017model, rusu2018meta}, optimization algorithm \cite{BaikCCKL20, zhang2021metanode}, embedding network \cite{snell2017prototypical, lee2019meta, zhang2020deepemd}, metric strategy \cite{Wertheimer_2021_CVPR, ZhangZNXY19, LiXHWGL19}, or label propagation strategy\cite{Tang_2021_CVPR, satorras2018few, distributionpropagation2020, rodriguez2020embedding}, etc. The representation learning-based approaches aim to design a good feature extractor \cite{YeHZS20, 0001FLWLH0X21} or training strategy \cite{Chen_2021_CVPR, tian2020rethinking, ChenLKWH19, Mangla0SKBK20} to learn transferable representations from abundant base classes, so that the novel class samples can be recognized by a simple cosine classifier \cite{chen2020new} or logistic regression classifier \cite{Rizve_2021_CVPR}. 

Recently, some studies attempt to introduce some external knowledge as auxiliary priors to further boost the performance \cite{boostingfew, xing2019adaptive, babysteps2019, zhimao2019few, zhang2021prototype} of existing FSL methods. For example, in \cite{xing2019adaptive, boostingfew}, the authors extend \cite{snell2017prototypical} by introducing the semantic information of class names as priors, to enhance class prototypes for FSL. In \cite{zhimao2019few}, they follow the representation learning-based approaches and introduce class hierarchies as priors to further boost cosine classifier \cite{chen2020new} for FSL. Different from these methods, our MetaDT focuses on interpretable FSL and introduces the priors of class hierarchy to construct and infer a novel decision tree for interpretable FSL. Its advantage is that more superclasses can be exploited to improve knowledge transfer and detailed decision explanations can be provided for class prediction. 

\begin{figure*}[t]
	\centering
	\includegraphics[width=1.0\textwidth]{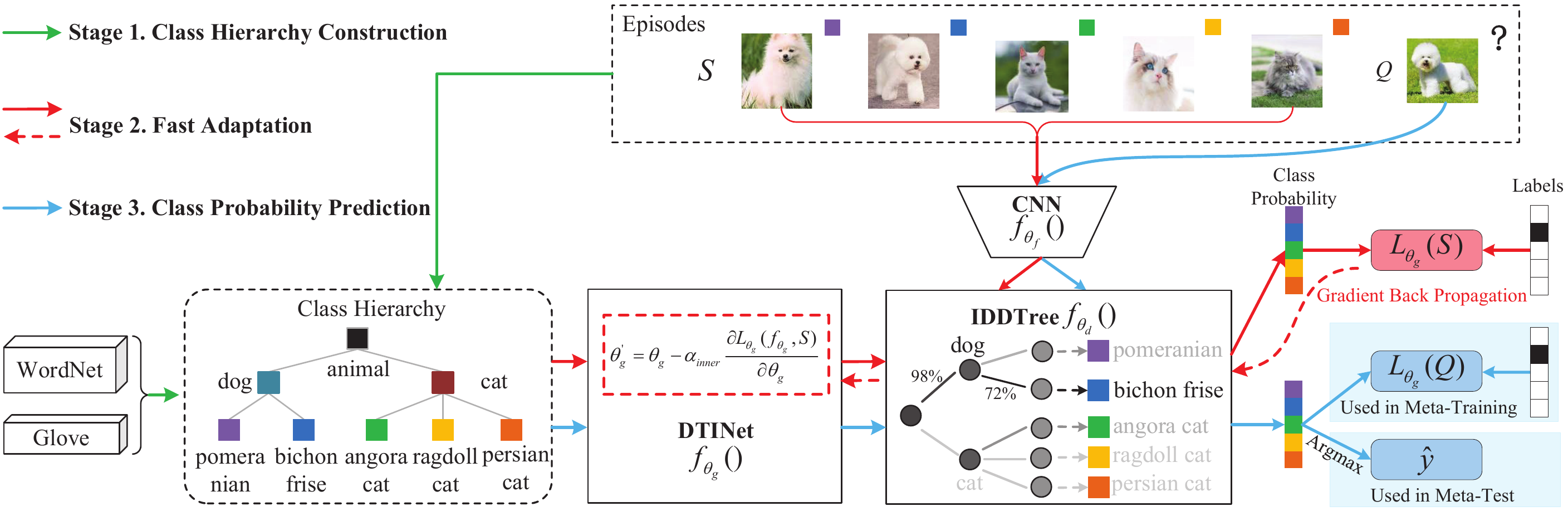} 
	\caption{The framework of our MetaDT. Given an episode, we first construct a class hierarchy as inputs of DTINet. Second, a task-specific IDDTree is inferred by finetuning the DTINet with $M$-step gradient updates on support set $\mathcal{S}$. Finally, the IDDTree is evaluated on query set $\mathcal{Q}$ and the evaluated results are used to train our DTINet in meta-training. In meta-test, the evaluated results are used to assign labels. 
	}
	\vspace{-15pt}
	\label{fig2}
\end{figure*} 

\subsection{Zero-Shot Learning}

Zero-Shot Learning (ZSL) is a challenging machine learning task, which targets at recognizing novel classes without any labeled examples by resorting to some semantic knowledge summarized from human's past experiences \cite{WanCLYZY019, FromeCSBDRM13}. The main idea is regarding class-level semantic knowledge as the auxiliary information, and then learning a map function between semantic knowledge and class representations for novel class prediction. The semantic knowledge can be class attributions \cite{WanCLYZY019}, class names \cite{xing2019adaptive}, class descriptions \cite{elhoseiny2019creativity}, or class hierarchies \cite{kampffmeyer2019rethinking}. Our work differs from these models in two aspects: 1) our MetaDT focuses on interpretable FSL, where few labeled samples should be fully exploited; 2) resorting to the semantic knowledge of class hierarchy, we focus on learning an interpretable decision tree, \emph{i.e.}, learning a map from class hierarchy to decision trees instead of learning class representations. 

\subsection{Interpretable Decision Models}
Decision model interpretability is a very popular research topic in machine learning, which focuses on clearing model decision process to ensure decision reliability \cite{carvalho2019machine}. In earlier studies \cite{safavian1991survey, lavanya2012ensemble}, the decision tree is a very popular model on a wide variety of tasks, as its attractive interpretability. After that, considering the high accuracy of neural networks, some works \cite{AlvinWan21, kontschieder2015deep, yang2018deep, KontschiederFCB16, YangSG19} further combine neural network and decision tree to improve accuracy and interpretability jointly.
Recently, several FSL studies also attempt to improve FSL interpretability \cite{CaoBL21, xue2020region, WangLVNKN21}. For example, in \cite{xue2020region}, Xue \emph{et al.} propose a region comparison network to highlight the region similarity between samples for the interpretable FSL. Wang \emph{et al.} \cite{WangLVNKN21} propose an attention-based explainable FSL model by highlighting the discriminative patterns of each image. Almost all methods improve FSL interpretability from the perspective of heatmap visualization, which does not directly reveal why FSL models make such decision. Different from these existing methods, we focus on the interpretability of FSL decision process and propose a novel decision tree-based meta-learning framework (\emph{i.e.}, MetaDT), which directly reveals the reason for the classification decision of each sample.  

\section{Problem Definition}
For the FSL, we are given two data sets: 1) a training set $\mathcal{S}$ (called support set) that contains $N$ novel classes $\mathcal{C}_{novel}$, with $K$ labeled samples per class where $K$ is very small (\emph{e.g.}, $K=1$ or 5); and 2) an auxiliary data set $\mathcal{D}_{base}$ that consists of abundant labeled images from base classes $\mathcal{C}_{base}$. Here, the sets of class $\mathcal{C}_{base}$ and $\mathcal{C}_{novel}$ are disjoint, \emph{i.e.}, $\mathcal{C}_{base} \cap \mathcal{C}_{novel} = \emptyset$. Based on the support set $\mathcal{S}$ and the auxiliary data set $\mathcal{D}_{base}$, our goal is to learn a good classifier for a test set $\mathcal{Q}$ (called query set) that consists of unlabeled novel class samples. This is a common setting in FSL studies, which is called $N$-way $K$-shot tasks (\emph{e.g.}, 5-way 1-shot tasks, or 5-way 5-shot tasks).

In this paper, we focus on the interpretable FSL. Our goal is to learn a decision-interpretable classifier for query set $\mathcal{Q}$, \emph{i.e.}, the learned classifier not only can perform novel class prediction but also can provide a clear explanation for making the class prediction. To this end, in this paper, we introduce a tree-like class hierarchy (i.e., the semantic hierarchy relations between classes) from external knowledge graphs (e.g., WordNet \cite{miller1998wordnet}) as FSL priors, and then carefully design a meta-learner leveraging the priors to effectively infer an interpretable decision tree for recognizing novel classes. 

\section{MetaDT Framework}
In this section, we introduce the technical details of the proposed MetaDT framework. As shown in Figure~\ref{fig2}, the framework consists of three key modules: a CNN-based feature extractor $f_{\theta_{f}}()$ with parameters $\theta_{f}$, a {d}ecision {t}ree {i}nference {net}work (DTINet) $f_{\theta_{g}}()$ with parameters $\theta_{g}$, and an {i}nterpretable and {d}ifferentiable {d}ecision {tree} (IDDTree) $f_{\theta_{d}}()$ with parameters $\theta_{d}$. Here, the feature extractor $f_{\theta_{f}}()$ aims to provide a good $d_f$-dimensional representation for each image, which is obtained by following the representation learning method \cite{Rizve_2021_CVPR}. The DTINet $f_{\theta_{g}}()$ is a meta-learner, which accounts for learning to quickly infer a task-specific decision tree $f_{\theta_{d}}()$ (\emph{i.e.}, IDDTree) by using the priors of class hierarchy and few labeled samples. The IDDTree $f_{\theta_{d}}()$ is a classifier where its parameters $\theta_{d}$ is not trainable but infered by the DTINet $f_{\theta_{g}}()$, which aims to perform class prediction for each support/query sample.  

\begin{figure}[t]
	\centering
	\includegraphics[width=0.47\textwidth]{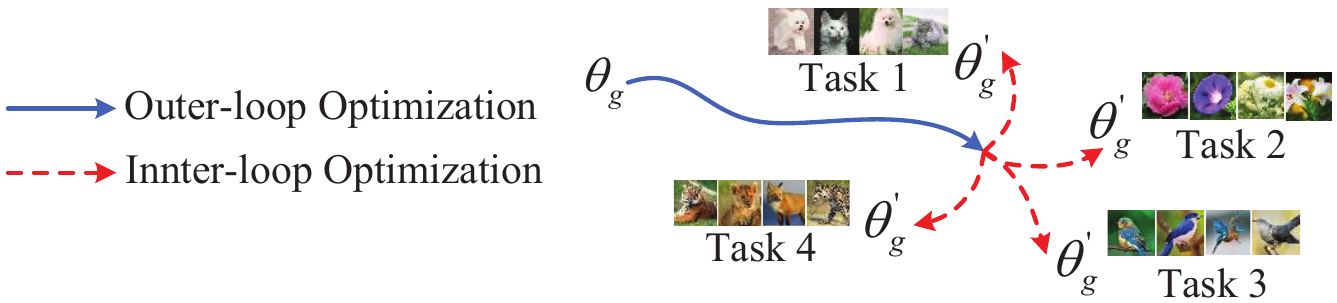} 
	\caption{The intuition for fast adaptation of decision tree.}
	\vspace{-15pt}
	\label{fig3}
\end{figure}

\subsection{Overall Workflow}
\label{section_3_1}
The main details of the IDDTree $f_{\theta_{d}}()$ and the DTINet $f_{\theta_{g}}()$ will be introduced in Sections~\ref{section_3_3} and \ref{section_3_4}, respectively. Here, we mainly present the workflow depicted in Figure~\ref{fig2}, including meta-training and meta-test phases.

\noindent {\bf Meta-Training.} The key challenge of our MetaDT is how to train our meta-learner (\emph{i.e.}, DTINet $f_{\theta_{g}}()$) by leveraging the priors of class hierarchy to quickly infer an interpretable decision tree (\emph{i.e.}, IDDTree $f_{\theta_{d}}()$) with very few labeled samples. To address the challenge, we employ a two-loop optimization mechanism \cite{finn2017model} (\emph{i.e.}, an inner-loop optimization and an outer-loop optimization) to learn the DTINet $f_{\theta_{g}}()$. As shown in Figure~\ref{fig3}, our intuition is treating the parameters $\theta_g$ of our meta-learner $f_{\theta_{g}}()$ as the initial parameters of inner-loop optimization, and then attempting to learn good (\emph{i.e.}, transferable) initial parameters $\theta_g$ by using an outer-loop optimization. By doing so, the initial model $f_{\theta_g}()$ can quickly adapt to novel classes with gradient updates upon few labeled samples $(x, y) \in \mathcal{S}$. 

Specifically, as shown in Figure~\ref{fig2}, following the episodic training manner \cite{vinyals2016matching}, we mimic the test setting and construct a number of $N$-way $K$-shot tasks (called episodes) from base classes. Then, we train our meta-learner $f_{\theta_{g}}()$ to quickly infer a task-specific IDDTree for each episode by minimizing the cross entropy loss on query set $\mathcal{Q}$. That is,
\begin{equation}
	\begin{aligned}
		&\min\limits_{\theta_g}\   \mathbb{E}_{(\mathcal{S},\mathcal{Q}) \in \mathbb{T}} L_{\theta_{g}}(\mathcal{Q}),\\
		L_{\theta_{g}}(\mathcal{Q})= \frac{1}{|\mathcal{Q}|} &\sum_{(x,y) \in \mathcal{Q}} CE(P(k|x, \mathcal{S}, \theta_{g}), y),
	\end{aligned}
	\label{eq0}
\end{equation}
where $\mathbb{T}$ is the set of constructed episodes and $CE()$ denotes the cross entropy loss function. The above training process can be regarded as an outer-loop optimization for learning good initial parameters $\theta_{g}$. Next, we introduce how to calculate the class probability $P(k|x, \mathcal{S}, \theta_g)$ that each query sample $x \in \mathcal{Q}$ belongs to class $k$ in an inner-loop optimization manner, given initial parameters $\theta_{g}$ and support set $\mathcal{S}$. It includes the following three stages:

{\bf \emph{Stage 1. Class Hierarchy Construction.}} We first construct a tree-like class hierarchy as FSL priors. The class hierarchy reflects the semantic relations between classes, \emph{i.e.}, what kinds of superclasses these few-shot classes share. For example, the classes ``pomeranian'' and ``bichon frise'' share the same superclass ``dog''. We note that such type of class hierarchy can be easily obtained from some external knowledge graphs, \emph{e.g.}, WordNet \cite{miller1998wordnet}. Then, we encode the tree-like class hierarchy by using an undirected graph $\mathcal{G}$, where its nodes denote all semantic classes (\emph{i.e.}, all few-shot classes and their superclasses) and its edges represent the hierarchy relations between all classes. Let $\mathcal{V}$ and $\mathcal{E}$ denote the set of nodes and edges respectively, \emph{i.e.}, $\mathcal{G}=(\mathcal{V}, \mathcal{E})$. In the graph $\mathcal{G}$, each node is represented by a $d_s$-dimensional semantic vector $h_i$, \emph{i.e.}, the mean of all $d_s$-dimensional Glove-based word embeddings \cite{pennington2014glove} of their class names, denoted by $\mathcal{H} = \{h_i\}_{i=0}^{F-1}$ where $F$ is the number of graph nodes. The edge set $\mathcal{E}$  is represented as a adjacency matrix $A \in \mathcal{R}^{F \times F}$ where $A_{i,j}=1$ if the node $i$ is associated with the node $j$, otherwise $A_{i,j}=0$. 

{\bf \emph{Stage 2. Fast Adaptation.}} In this stage, we regard the class hierarchy $\mathcal{G}$ as inputs and the IDDTree $f_{\theta_{d}}()$ as predicted targets of our initial DTINet $f_{\theta_{g}}()$. Then, we evaluate the predicted IDDTree $f_{\theta_{d}}()$ on the support set $\mathcal{S}$ and leverage the evaluated classification loss $L_{\theta_{g}}(\mathcal{S})$ to quickly finetune the initial DTINet $f_{\theta_{g}}()$ with $M$-step gradient updates. As a result, a task-specific DTINet $f_{\theta'_{g}}()$ with parameters $\theta'_{g}$ can be obtained for each few-shot task. For example, when using one gradient updates (\emph{i.e.}, $M$=1), the process of such fast adaptation can be expressed as:
\begin{equation} 
	\begin{aligned}
	\theta_{g}^{'} &= \theta_{g} - \alpha_{inner} \triangledown L_{\theta_{g}}(\mathcal{S}),
	\end{aligned}
	\label{eq1}
\end{equation}
where $\alpha_{inner}$ is the learning rate of gradient updates. The above process can be regarded as an inner-loop optimization for learning good task-specific parameters $\theta'_{g}$. Next, we elaborate on the calculation details for the classification loss $L_{\theta_{g}}(\mathcal{S})$ on the support set $\mathcal{S}$.

Speficifally, we first extract the features of each support sample $(x, y) \in \mathcal{S}$ via the feature extractor $f_{\theta_{f}}()$. Then, the class probability $P(k|x, \theta_{g})$ that  each sample $(x, y) \in \mathcal{S}$ belongs to class $k$ can be calculated by using the IDDTree $f_{\theta_{d}}()$ infered by our meta-learner $f_{\theta_g}()$ (Please refers to Section~\ref{section_3_3} for its calculation details). That is, 
\begin{equation} 
	\begin{aligned}
		&\theta_{d} = f_{\theta_{g}}(\mathcal{G}), \\
		P(k|x, \theta_{g})& = f_{\theta_{d}}(f_{\theta_{f}}(x)),\ (x,y) \in \mathcal{S}.
	\end{aligned}
	\label{eq2}
\end{equation}
Finally, we calculate the classification loss $L_{\theta_{g}}(f_{\theta_{g}}, \mathcal{S})$ by using a cross entropy loss on support set $\mathcal{S}$:
\begin{equation} 
	\begin{aligned}
		L_{\theta_{g}}(\mathcal{S})=\frac{1}{|\mathcal{S}|}\sum_{(x,y) \in \mathcal{S}} CE(P(k|x, &\theta_{g}), y).
	\end{aligned}
	\label{eq3}
\end{equation}

{\bf \emph{Stage 3. Class Probability Prediction.}} Based on the above task-specific DTINet $f_{\theta_{g}^{'}}()$, we can obtain a task-specific IDDTree $f_{\theta'_d}()$. Then, the class probability $P(k|x, \mathcal{S}, \theta_{g})$ that each query sample $x \in \mathcal{Q}$ belongs to class $k$ can be calculated in a sequential decision manner (Please refers to Section~\ref{section_3_3} for its details). That is,
\begin{equation} 
	\begin{aligned}
		&\theta'_{d} = f_{\theta_{g}^{'}}(\mathcal{G}), \\
		P(k|x, &\mathcal{S}, \theta_{g}) = f_{\theta'_{d}}(x), x \in \mathcal{Q}.
	\end{aligned}
	\label{eq4}
\end{equation}

\noindent {\bf Meta-Test.} The workflow of meta-test is similar to meta-training. As shown in Figure~\ref{fig2}, the difference is that we remove the training step of Eq.~\ref{eq0} and directly evaluate the class probability $P(k|x, \mathcal{S}, \theta_{g})$ of each query sample $x \in \mathcal{Q}$ by following Eqs.~\ref{eq1} $\sim$ \ref{eq4}. Finally, we perform the novel class prediction $\hat{y}$ to the novel class with highest probability:
\begin{equation} 
		\hat{y}=\mathop{\arg\max}_{k \in [0,N-1]}P(k|x, \mathcal{S}, \theta_{g}).
	\label{eq4-1}
\end{equation}

\subsection{Interpretable and Differentiable Decision Tree}
\label{section_3_3}

In the subsection, we introduce the technical details of our IDDTree $f_{\theta_{d}}()$, including the following two parts.

\begin{figure}[t]
	\centering
	\includegraphics[width=0.475\textwidth]{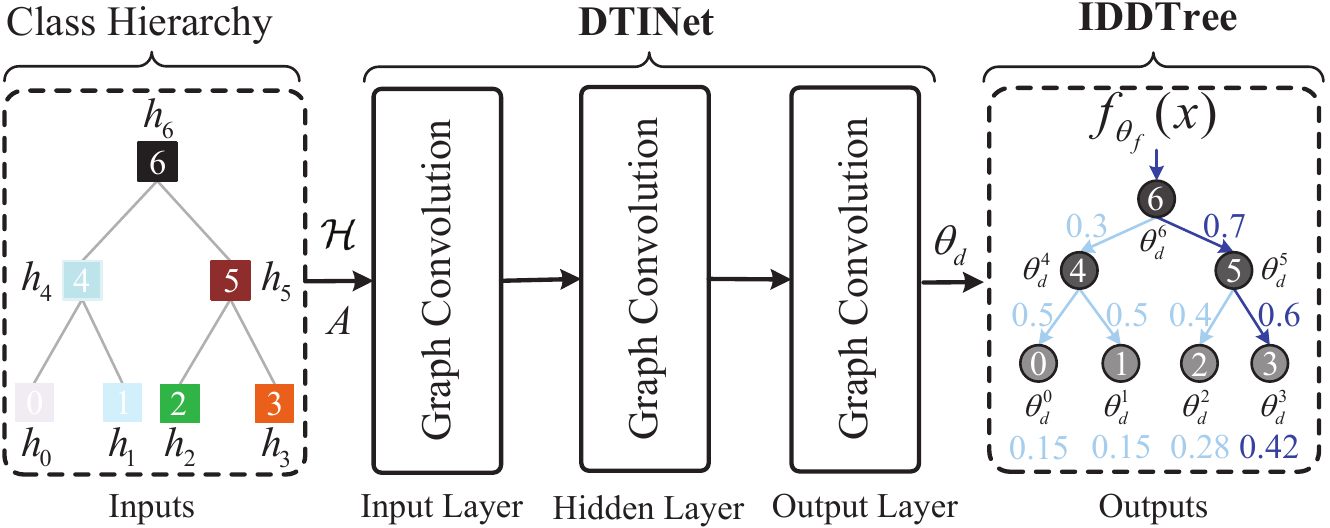} 
	\caption{Details of DTINet and IDDTree. $\mathcal{H}$ and $A$ is the sementic vectors and adjacency matrix of class hierarchy, respectively. $\theta_{d}$ is the outputs where $\theta^{i}_{d}$ is correspondings to $h_i$, $i=0,1,...,6$.}
	\vspace{-15pt}
	\label{fig4}
\end{figure}

\noindent {\bf How to design the IDDTree $f_{\theta_{d}}()$?} As shown in Figure~\ref{fig4}, for interpretability, our main idea is treating the hierarchy relations in the tree-like class hierarchy $\mathcal{G}$ as explicit decision rules, and then performing class predictions by following the rules. 
For differentiability, our notion is assigning each tree node $i$ of the class hierarchy $\mathcal{G}$ with a $d_f$-dimensional weight vector $\theta^{i}_{d} \in \mathcal{R}^{d_f}$ (\emph{i.e.}, $\theta_{d}=\{\theta^{i}_{d}\}_{i=0}^{F-1}$ where $F$ is the number of tree nodes), and then performing the class prediction in a sequential and soft probability inference manner from the root of decision rules to its leafs. 

\noindent {\bf How to evaluate the class probability?} Intuitively, the weight vector $\theta^{i}_{d}$ can be regarded as the prototype or class center of each class $i \in [0, F-1]$ in the feature space. Thus, given a support/query image $x \in \mathcal{S}/\mathcal{Q}$ and its features $f_{\theta_{f}}(x)$, we can evaluate the class probability $P(k|x, \theta_{d})$ that the sample $x$ belongs to few-shot class $k \in [0, N-1]$ in a metric-based decision manner from the root of decision tree to its leaf nodes. It includes the following two steps: 

{\bf\emph{Step 1.}} For each non-leaf node $i$ (\emph{i.e.}, superclasses), we need to make decisions for its all child nodes denoted by $\mathcal{C}_{child}^{i}$, \emph{i.e.}, evaluating the conditional probability $P(j|i, x, \theta_{d})$ that the sample $x$ belongs to child class $j \in \mathcal{C}_{child}^{i}$. 
We evaluate the conditional probability $P(j|i, x, \theta_{d})$ by computing the cosine similarity between the sample $x$ and the weights $\theta^{j}_{d}$ of child class $j \in \mathcal{C}_{child}^{i}$. That is,
\begin{equation}
	P(j|i, x, \theta_{d}) = \frac{e^{<f_{\theta_f}(x),\ \theta^{j}_{d}>\ \cdot\ \gamma}}
	{\sum_{c \in \mathcal{C}_{child}^{i}} e^{<f_{\theta_f}(x),\ \theta^{c}_{d}>\ \cdot\ \gamma}},
	\label{eq5}
\end{equation}
where $<\cdot>$ denotes the cosine similarity of two vectors and $\gamma$ is a scale parameter. Following \cite{ChenLKWH19}, $\gamma=10$ is used.

{\bf\emph{Step 2.}} For each leaf node (\emph{i.e.}, few-shot class $k$), given its traversal path $\tau_k$ from the root and the conditional probability $P(j|i, x, \theta_{d})$ of each node $i \in \tau_k$ traversing its child node $j$ in the path $\tau_k$, \emph{i.e.}, $j \in 
\tau_k \cap \mathcal{C}_{child}^{i}$, the class probability $P(k|x, \theta_{d})$ that each support/query sample $x$ belongs to the class $k$ can be calculated in a cumulative product manner along the traversal path $\tau_k$. That is,
\begin{equation}
	P(k|x, \theta_{d}) = \prod \limits_{i \in \tau_k} P(j|i,x,\theta_{d}), j \in \tau_k \cap \mathcal{C}_{child}^{i}.
	\label{eq6}
\end{equation}
For clarity, we give an example to show the above calculation process. As shown in Figure~\ref{fig4}, given a sample $x$, 1) we compute all conditional probability of non-leaf nodes in Eq.~\ref{eq5}, \emph{e.g.}, $P(5|6, x, \theta_{d}) = 0.7$; and then 2) the class probability is obtained by using Eq.~\ref{eq6}, \emph{e.g.}, $P(3|x, \theta_{d}) = P(6|x, \theta_{d}) P(5|6, x, \theta_{d})P(3|5, x, \theta_{d}) = 1 \cdot 0.7 \cdot 0.6 = 0.42$. 

\subsection{Decision Tree Inference Network}
\label{section_3_4}
In the IDDTree $f_{\theta_{d}}()$, its key challenge is how to quickly infer its parameters $\theta_{d}=\{\theta^{i}_{d}\}_{i=0}^{F-1}$ when only few labeled samples are available.  To address the challenge, we regard the class hierarchy $\mathcal{G}$ as inputs and then design a graph convolution-based inference network $f_{\theta_{g}}()$ to learn the map from class hierarchy $\mathcal{G}$ to the parameters $\theta_{d}$. The intuition behind this design is fully leveraging human's prior knowledge to enable fast adaptation of decision trees. 

The overall structure is shown in Figure~\ref{fig4}, which consists of an input layer, a hidden layer, and an output layer. The input and hidden layers aim to obtain a good representation for each graph node by exploiting abundant semantic information and hierarchy relations between classes. Then, the parameters $\theta_{d}$ of IDDTree $f_{\theta_{d}}()$ are predicted by the final output layer. The overall process can be defined as:
\begin{equation}
	\begin{aligned}
		\theta_{d} = f_{\theta_{g}}(\mathcal{H}, A) = \hat{A}\delta(\hat{A}\delta(\hat{A}\mathcal{H}W^{(0)})W^{(1)})W^{(2)}
	\end{aligned}
	\label{eq7}
\end{equation}
where $W^{(0)}$, $W^{(1)}$, and $W^{(2)}$ denote a trainable weight matrix for input, hidden and output layers, respectively, \emph{i.e.}, $\theta_{g}=\{W^{(0)}, W^{(1)}, W^{(2)}\}$; $\hat{A}$ is the normalized $A$, \emph{i.e.}, $\hat{A}=D^{-\frac{1}{2}}AD^{\frac{1}{2}}$; $D$ is the degree matrix, \emph{i.e.}, $D_{ii} = \sum_{j} A_{ij}$; and $\delta()$ denotes the ReLU activation function.

\begin{table*}[t]
	\caption{Comparison with state-of-the-art methods on miniImagenet and CIFAR-FS. The best results are highlighted in bold.}\smallskip
	\centering
	\smallskip\scalebox
	{0.85}{\begin{tabular}{l|c|c|c|c|c|c}
			\hline
			\multicolumn{1}{l|}{\multirow{2}{*}{Method}}&
			\multicolumn{1}{c|}{\multirow{2}{*}{Using Knowledge}}& \multicolumn{1}{c|}{\multirow{2}{*}{Backbone}}&
			\multicolumn{2}{|c|}{miniImagenet} & \multicolumn{2}{|c}{CIFAR-FS} \\ 
			\cline{4-7}
			& & & 5-way 1-shot & 5-way 5-shot & 5-way 1-shot & 5-way 5-shot \\
			\hline \hline
			CGCS \cite{gao2021curvature} & No & ResNet12 & $67.02 \pm 0.20\%$  & $82.32 \pm 0.14\%$ & $71.66 \pm 0.23\%$  & $ 85.50 \pm 0.15\%$ \\ 
			RCNet \cite{xue2020region} & No & ResNet12 &  $ 57.40 \pm  0.86\%$  & $75.19 \pm 0.64\%$ & $69.02 \pm 0.92\%$  & $82.96 \pm 0.67\%$ \\
			ALFA \cite{BaikCCKL20} & No & ResNet12 & $59.74 \pm 0.49\%$  & $77.96 \pm 0.41\%$ & -  & - \\
			MeTAL \cite{baik2021meta} & No & ResNet12 & $59.64 \pm 0.38\%$  & $76.20 \pm 0.19\%$ & $67.97 \pm 0.47\%$  & $82.17 \pm 0.38\%$ \\
			CRF-GNN \cite{Tang_2021_CVPR} & No & ConvNet256 &  57.89 $\pm$ 0.87$\%$  & 73.58 $\pm$ 0.87$\%$ & 76.45 $\pm$ 0.99$\%$  & 88.42 $\pm$ 0.23$\%$ \\
			Neg-Cosine \cite{liu2020negative} & No & ResNet12 &  $63.85 \pm 0.81\%$  & 81.57 $\pm$ 0.56\% & -  & - \\
			RFS \cite{tian2020rethinking} & No & ResNet12 &  $62.02 \pm 0.63\%$  & 79.64 $\pm$ 0.44\% & $71.51 \pm 0.80\%$  & $86.00 \pm 0.50\%$ \\
			InvEq \cite{Rizve_2021_CVPR} & No & ResNet12 &  $66.82\pm  0.80\%$  & $84.35 \pm 0.51\%$ & $76.83 \pm 0.82\%$  & $89.26 \pm 0.58\%$ \\
			
			TriNet \cite{ChenFZJXS19} & Yes & ResNet18 &  $58.12 \pm 1.37\%$  & $76.92 \pm 0.69\%$ & -  & - \\
			AM3-PNet \cite{xing2019adaptive} & Yes & ResNet12 &  $65.21 \pm 0.30\%$  & $75.20 \pm 0.27\%$ & -  & - \\
			AM3-TRAML \cite{boostingfew} & Yes & ResNet12 &  67.10 $\pm$ 0.52$\%$  & 79.54 $\pm$ 0.60\% & -  & - \\
			FSLKT \cite{zhimao2019few} & Yes & ConvNet128 & $64.42 \pm 0.72\%$  & $74.16 \pm 0.56\%$ & -  & - \\
			
			\cline{1-7}
			MetaDT & Yes & ResNet12 & 69.08 $\pm$ 0.73$\%$ & 83.40 $\pm$ 0.51$\%$ & 79.03 $\pm$ 0.75$\%$ & 88.50 $\pm$ 0.58$\%$ \\
			MetaDT + Cosine Classifier & Yes & ResNet12 & \textbf{70.45} $\pm$ \textbf{0.81}$\%$ & \textbf{84.84} $\pm$ \textbf{0.52}$\%$ & \textbf{80.14} $\pm$ \textbf{0.78}$\%$ & \textbf{89.84} $\pm$ \textbf{0.56}$\%$ \\
			\hline
	\end{tabular}}
	\label{table1}
\end{table*}

\section{Enhancement: MetaDT + Cosine Classifier}
\label{section_3_6}
Till now, we have introduced all details of our MetaDT. Its advantage is that the decision process of prediction is interpretable resorting to the priors of class hierarchy. In this section, we introduce how to fuse our MetaDT and the existing cosine classifier \cite{chen2020new} for better performance. 

Intuitively, the two methods are complementary to each other: 1) when the labeled samples are very scarce (\emph{e.g.}, $K$=1), our MetaDT is more reliable because it fully exploits the priors of class hierarchy which effectively alleviates the data sparsity issue; and 2) as more and more labeled samples become available, the data sparsity issue gradually disappears and the existing cosine classifier becomes more effective. Thus, we propose to fuse the two methods in class probability via a convex combination manner. That is,
\begin{small}
\begin{equation}
	\begin{aligned}
		P_{fused}(k|x) = \lambda P(k|x, \mathcal{S}, \theta_{g}) + (1-\lambda)P_{cos}(k|x, \mathcal{S}),
	\end{aligned}
	\label{eq9}
\end{equation}
\end{small}where $P(k|x, \mathcal{S}, \theta_{g})$, $P_{cos}(k|x, \mathcal{S})$, and $P_{fused}(k|x)$ denote the class probability of our MetaDT, the cosine classifier, and after fusion, respectively; $\lambda \in [0, 1]$ is a weight parameter. Finally, we perform the class prediction of query sample $x \in \mathcal{Q}$ by using fused probability $P_{fused}(k|x)$. The fusion strategy is only used in meta-test. In real applications, the ``MetaDT + Cosine Classifier'' can be used when we merely go for better performance; otherwise our MetaDT would be a good choice for FSL, which not only delivers promising performance but also has clear interpretability. 
\section{Experiments}

\subsection{Datasets and Settings}
\noindent \textbf{miniImagenet.} The dataset is a subset from ImageNet, containing 100 classes. Following \cite{vinyals2016matching}, we split it into 64, 16, and 20 classes for training, validation, and test, respectively. Besides, the tree-like class hierarchy is constructed from WordNet by using the relation of ``hypernyms()''. 

\noindent \textbf{CIFAR-FS.}
The dataset is built from CIFAR100, including 100 classes. Following \cite{BertinettoHTV19}, we split the data set into 64 classes for training, 16 classes for validation, and 20 classes for test, respectively. Similarly, its tree-like class hierarchy is also acquired from WordNet via its class names.

\noindent \textbf{tieredImagenet.} 
The dataset is a more challenging FSL dataset constructed from ImageNet, including 608 classes. Following \cite{RenTRSSTLZ18}, we partition the dataset into 34 high-level classes, and then split it into 20 classes for training, 6 classes for validation, and 8 classes for test. Similarly, its class hierarchy is also extracted from WordNet. 

\subsection{Implementation Details}
\noindent \textbf{Architecture.}
We employ the ResNet12 trained following previous work \cite{Rizve_2021_CVPR} as our feature extractor. For DTINet, we use a three-layer graph convolution with a 300-dimensional input, a 1024-dimensional input layer, a 2048-dimensional hidden layer, and a 640-dimensional output layer to infer the IDDTree. In each layer, ReLU is used as the activation function except for the output layer, and a dropout layer with probability of 0.5 is used for better generalization.

\noindent \textbf{Training and Test Details.}
In meta-training phase, we train our meta-learner with 20 epochs by using an Adam optimizer with a weight decay of 0.0005 and a learning rate of 0.0001. 
The update steps $M$ and learning rate $\alpha_{inner}$ of inner-loop optimization is set to 25 and 0.05, respectively. In meta-test phase, the learning rate $\alpha_{inner}$ remains unchanged, and the update step $M$ is changed to 50, 70, and 150 on miniImagenet, CIFAR-FS, and tieredImagenet datasets, respectively. For the weight parameter $\lambda$, we set it to 0.8 and 0.1 for 1-shot and 5-shot tasks, respectively.


\noindent \textbf{Evaluation.} We evaluate our methods on 600 5-way 1-shot/5-shot tasks randomly sampled from the test set. 
Lastly, the mean accuracy together with the 95\% confidence interval is reported as the evaluation results.

\subsection{Experimental Results}

\begin{table}[t]
	\caption{Comparison with state-of-the-art methods on tieredImagenet. 
		Here, ``CC'' denotes ``Cosine Classifier''. 
	}\smallskip
	\centering
	\smallskip\scalebox
	{0.80}{\begin{tabular}{l|c|c|c}
			\hline
			\multicolumn{1}{l|}{\multirow{2}{*}{Method}} & \multicolumn{1}{l|}{\multirow{2}{*}{Backbobe}} & \multicolumn{2}{|c}{tieredImagenet}\\ 
			\cline{3-4}
			& & 5-way 1-shot & 5-way 5-shot\\
			\hline \hline
			CGCS \cite{gao2021curvature} & ResNet12 & $71.66 \pm 0.23\%$  & $ 85.50 \pm 0.15\%$ \\ 
			ALFA \cite{BaikCCKL20} & ResNet12 & $64.62 \pm 0.49\%$  & $82.48 \pm 0.38\%$ \\
			MeTAL \cite{baik2021meta} & ResNet12 & $63.89 \pm 0.43\%$  & $80.14 \pm 0.40\%$ \\
			CRF-GNN \cite{Tang_2021_CVPR} & ConvNet256 & 58.45 $\pm$ 0.59$\%$  & 74.58 $\pm$ 0.84$\%$ \\
			RFS \cite{tian2020rethinking} & ResNet12 & $69.74 \pm 0.72\%$  & 84.41 $\pm$ 0.55\% \\
			InvEq \cite{Rizve_2021_CVPR} & ResNet12 & $71.87 \pm 0.89\%$  & $86.82 \pm 0.58\%$ \\
			
			AM3-PNet \cite{xing2019adaptive} & ResNet12 & $67.23 \pm 0.34\%$  & $78.95 \pm 0.22\%$ \\
			
			\cline{1-4}
			MetaDT & ResNet12 & 70.56 $\pm$ 0.90$\%$ & 85.17 $\pm$ 0.56$\%$ \\
			MetaDT + CC & ResNet12 & \textbf{72.69} $\pm$ \textbf{0.90}$\%$ & \textbf{87.10} $\pm$ \textbf{0.56}$\%$ \\
			\hline
	\end{tabular}}
	\label{table2}
\end{table}

\begin{table}
	\centering
	\caption{Percentage of seen and unseen superclasses. }\smallskip
	\smallskip\scalebox
	{0.80}{
		\smallskip\begin{tabular}{l|c|c|c}
			\hline
			Types & miniImagenet & CIFAR-FS & tieredImagenet \\
			\hline \hline
			Seen Superclasses & 97.30\% & 85\% & 49.60\% \\
			Unseen Superclasses & 2.70\% & 15.00\% & 50.40\% \\
			\hline
	\end{tabular}}
	\label{table3}
\end{table}

\noindent {\bf Results on miniImagenet and CIFAR-FS.} Table~\ref{table1} shows the experimental results of various methods on miniImagenet and CIFAR-FS. From these results, we observe that our ``MetaDT + Cosine Classifier'' exceeds these state-of-the-art methods by around 1\% $\sim$ 4\%, which shows the superiority of our method on FSL. In addition, it can be found that our MetaDT achieves superior performance over existing methods by around 2\% $\sim$ 3\% on 1-shot tasks, as well as comparable performance on 5-shot tasks. Note that our MetaDT is decision-interpretable. Specifically, compared with these methods without exploiting knowledge, our MetaDT additionally explores the priors of class hierarchy, and mainly focuses on leveraging these priors to quickly infer a decision tree. The results validate the effectiveness of our decision tree classifier. It is worth noting that RCNet \cite{xue2020region} is also developed for interpretable FSL, which is a strong competitor. Our MetaDT significantly outperforms RCNet by a large margin by 6\% $\sim$ 12\%.
This further shows the superiority of our MetaDT for interpretable FSL. 

As for the FSL methods using knowledge, they also explore the semantic knowledge as priors for improving FSL but mainly use the nearest neighbor classifier. Different from these methods, our MetaDT leverages the priors of class hierarchy to quickly infer a task-specific decision tree for interpretable FSL. The results demonstrate the effectiveness of our MetaDT. Note that our MetaDT exceeds FSLKT around 6\% $\sim$ 10\%, which also utilizes the priors of class hierarchy. Different from FSLKT, our MetaDT regards the class hierarchy as a decision tree, so that more superclasses can be exploited for novel class prediction. The results show the superiority of our MetaDT to leverage class hierarchy. 

\noindent {\bf Results on tieredImagenet.} In Table~\ref{table2}, we report the classification results of our methods and various baseline methods on tieredImagenet. We find that our ``MetaDT + Cosine Classifier'' achieves superior performance over state-of-the-art methods. In addition, our MetaDT achieves competitive performance with some state-of-the-art methods. It is worth noting that the performance improvement of our MetaDT on tieredImagenet is limited, which is inconsistent with the one of miniImagenet and CIFAR-FS. This may be reasonable because the tieredImagenet dataset is split by following the high-level semantic classes. As a result, some superclasses are unseen in our meta-training phase, which impedes the knowledge transfer of our MetaDT. 

To illustrate this point, we calculate the percentage of seen and unseen superclasses on miniImagenet, CIFAR-FS, and tieredImagenet, in Table~$\ref{table3}$. From the results, we indeed find that the percentage of unseen superclasses of tieredImagenet is much larger than the one of miniImagenet and CIFAR-FS. This means the meta knowledge learned from the base classes is difficult to transfer to the novel classes. For the issue, we will leave it to future work.

\subsection{Ablation Study}
We conduct various experiments on miniImagenet to investigate the impacts of different components. 

\begin{table}
	\centering
	\caption{Ablation study of different components for our MetaDT. }\smallskip
	\smallskip\scalebox
	{0.82}{
		\smallskip\begin{tabular}{l|c|c|c}
			\hline
			 & Settings & 5-way 1-shot & 5-way 5-shot \\
			\hline \hline
			(\romannumeral1) & MetaDT & $69.08 \pm 0.73\%$ & $83.40 \pm 0.51\%$\\
			(\romannumeral2) & w/o Class Semantic & $67.23 \pm 0.83\%$ & $81.21 \pm 0.54\%$ \\
			(\romannumeral3) & w/o Graph Convolution & $66.79 \pm 0.73\%$ & $82.18 \pm 0.52\%$ \\
			(\romannumeral4) & w/o DTINet & $67.14 \pm 0.73\%$ & $81.12 \pm 0.54\%$ \\
			
			(\romannumeral5) & w/o Fast Adaptation & $57.04 \pm 0.86\%$ & $57.07 \pm 0.83\%$ \\
			\hline
	\end{tabular}}
	\label{table4}
\end{table}

\begin{table}
	\centering
	\caption{Ablation study of the proposed fusion strategy. 
	}\smallskip
	\smallskip\scalebox
	{0.78}{
		\smallskip\begin{tabular}{l|c|c|c}
			\hline
			& Settings & 5-way 1-shot & 5-way 5-shot \\
			\hline \hline
			\multicolumn{1}{l|}{\multirow{2}{*}{(\romannumeral1)}} & MetaDT & $69.08 \pm 0.73\%$ & $83.40 \pm 0.51\%$\\
			& Cosine Classifier & $66.96 \pm 0.81\%$ & $84.53 \pm 0.52\%$\\
			\hline
			\multicolumn{1}{l|}{\multirow{1}{*}{(\romannumeral2)}} & MetaDT + Cosine Classifier & $70.45 \pm 0.81\%$ & $84.84 \pm 0.52\%$ \\
			\hline
	\end{tabular}}
	\label{table5}
\end{table}

\begin{figure}
	\centering
	\includegraphics[width=0.85\columnwidth]{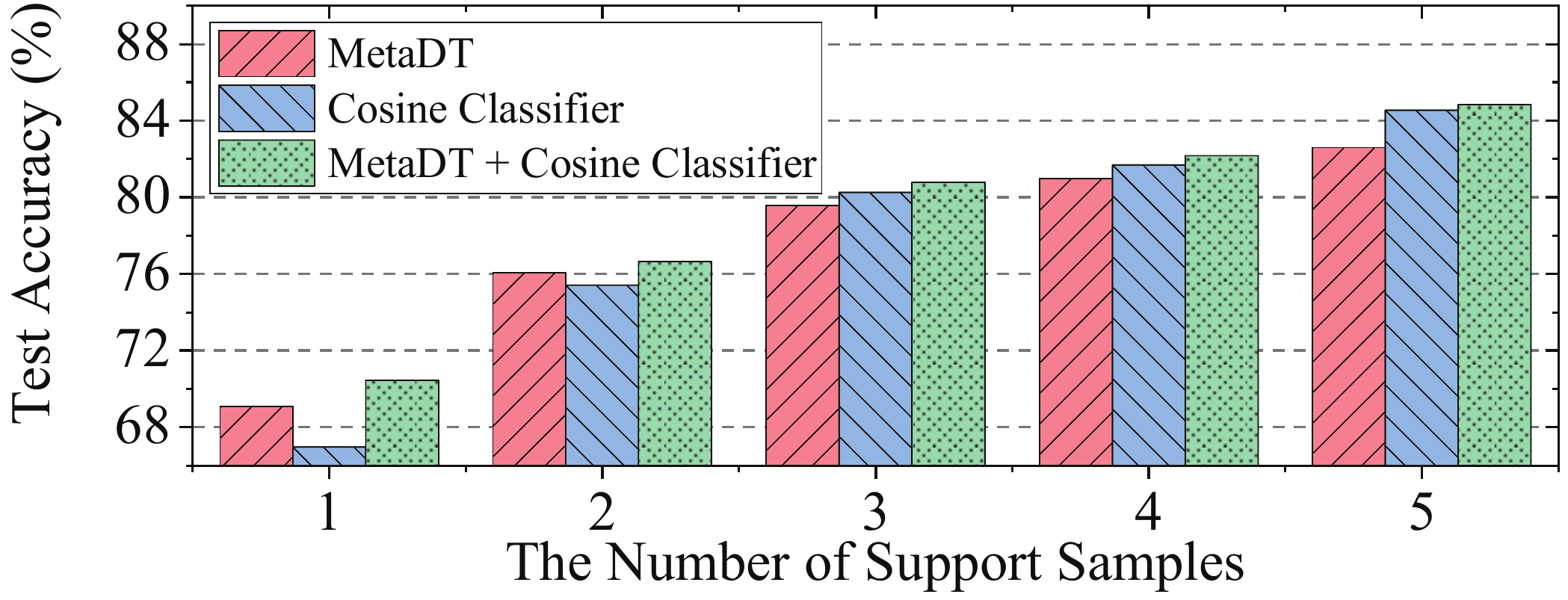}
	\caption{Analysis of 5-way $K$-shot tasks with different $K$ values.}
	\label{fig5}
\end{figure}

\noindent \textbf{How do different components affect our MetaDT?} We conduct various ablation studies to analyze the contribution of different components, including class semantic, class hierarchy, graph convolution, DTINet, and fast adaptation. Specifically, (\romannumeral1) we evaluate our MetaDT on 5-way 1/5-shot tasks; (\romannumeral2) we remove the semantic vectors $\mathcal{H}$ and replace it by using one-hot vectors on (\romannumeral1); (\romannumeral3) we replace the three-layer graph convolution by using a three-layer fully-connected network on (\romannumeral1); 
(\romannumeral4) we remove the DTINet on (i) and estimate the parameters of IDDTree in an average manner on mean-based prototypes by following the class hierarchy relations; 
and (\romannumeral5) we remove the fast adaptation of our meta-learner on (\romannumeral1). These results are shown in Table~\ref{table4}. From these results, we can see that the classification performance decreases by around 2\% $\sim$ 16\% when removing these components respectively. This suggests that employing these four key components is beneficial for our MetaDT. 

\noindent \textbf{Is the fusion of our MetaDT and cosine classifier effective?} In Table~\ref{table5}, we analyze the effectiveness of the fusion strategy described in Section~\ref{section_3_6}. Specifically, (\romannumeral1) we evaluate our MetaDT and cosine classifier, respectively; (\romannumeral2) following Eq.~\ref{eq9}, we fuse our MetaDT with cosine classifier. From the results, we find that the performance of our ``MetaDT + Cosine Classifier'' exceeds our MetaDT and the cosine classifier by around 1\% $\sim$ 4\%. This shows the superiority of the proposed fusion strategy on improving FSL performance. 

\begin{figure}
	\centering
	\begin{subfigure}{0.46\linewidth}
		\includegraphics[width=1.0\columnwidth]{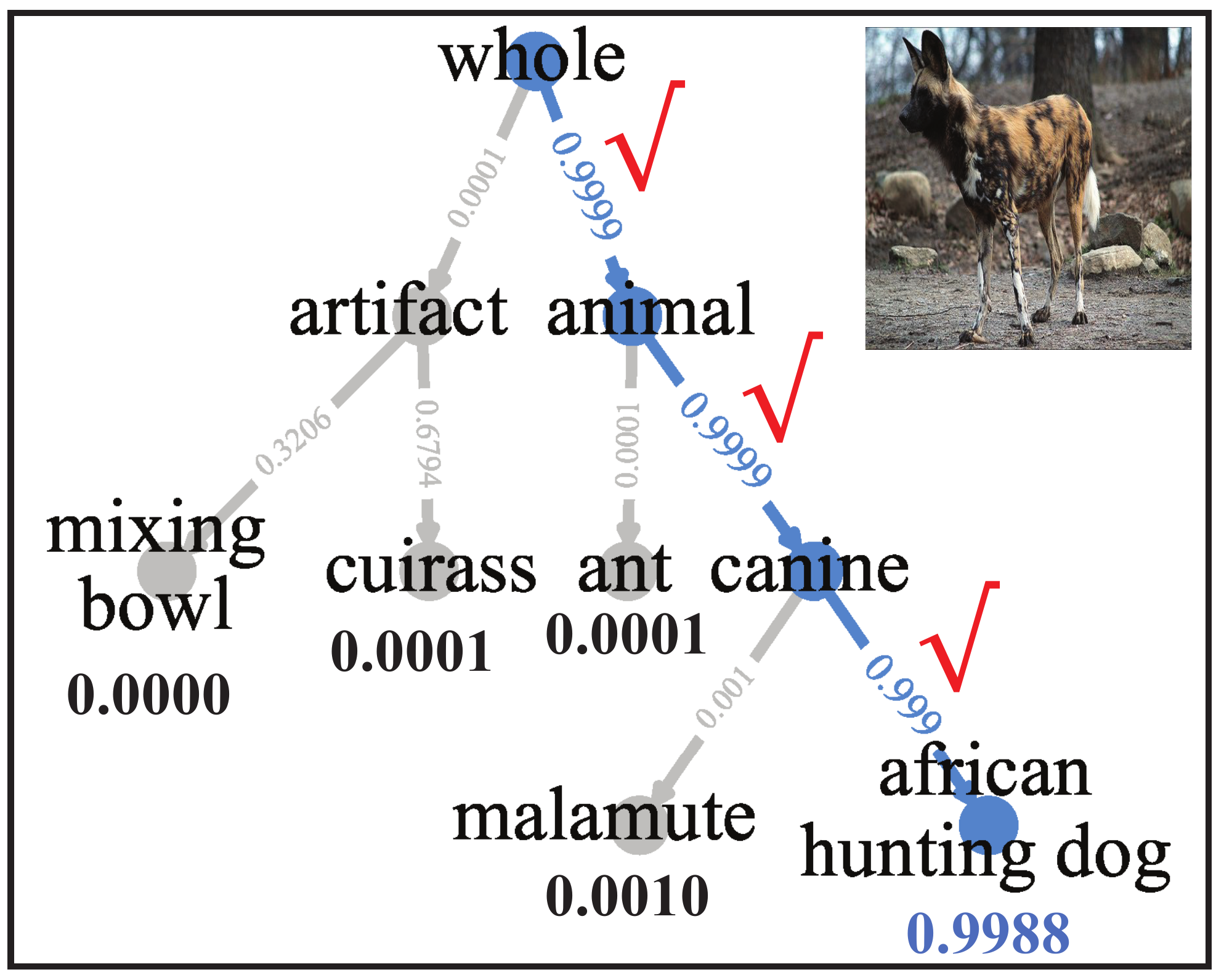}
		\caption{A case of right decision.}
		\label{fig6a}
	\end{subfigure}
	\hfill
	\begin{subfigure}{0.46\linewidth}
		\includegraphics[width=1.0\columnwidth]{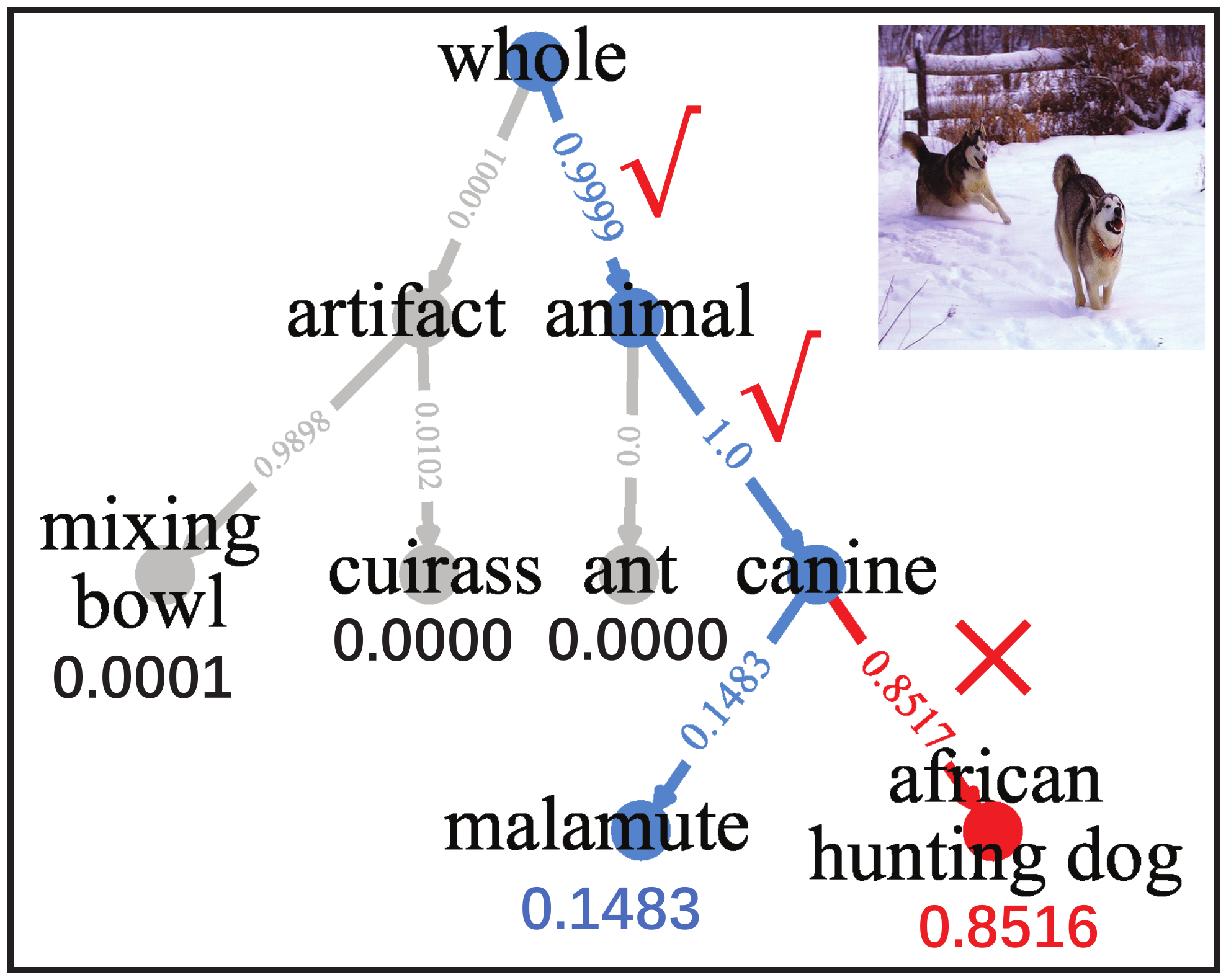}
		\caption{A case of wrong decision.}
		\label{fig6b}
	\end{subfigure}
	\caption{Decision interpretability analysis of our MetaDT.}
	\label{fig6}
\end{figure}

\noindent \textbf{How does our method perform under different numbers of support samples?} In Figure~\ref{fig5}, we count the performance of our MetaDT, cosine classifier, and our ``MetaDT + Cosine Classifier'' on different numbers of support samples. We find that 1) our MetaDT is more reliable on 1/2-shot tasks while the cosine classifier is more effective on 3/4/5-shot tasks; 2) our ``MetaDT + Cosine Classifier'' performs best. This verifies that our MetaDT and the cosine classifier indeed remedy the shortcomings of each other. 

\subsection{Interpretability and Visualization Analysis}
\noindent \textbf{Is the decision of our MetaDT interpretable?} To show the decision interpretability of our MetaDT, we conduct two 5-way 1-shot case studies on miniImagenet, including a right and a wrong decision case. Specifically, we randomly select a 5-way 1-shot task from test set. Then we construct and learn a four-layer decision tree by using only one labeled sample. After that, we randomly select an image that is rightly predicted and an image that is wrongly predicted as an example to illustrate the interpretable decision process, which are shown in Figure~\ref{fig6}. From Figure~\ref{fig6a}, we can see that our decision tree makes right decision in entire sequential decision path for the image of ``african hunting dog'', where all decisions have real meaning and fit our intuition. Besides, we note that such interpretable decison process can also help us trace and locate the root reason of making such wrong decisions. As shown in Figure~\ref{fig6b}, the reason causing the image of ``malamute'' is misidentified is that the image is misclassified as ``african hunting dog'' when performing the decision of ``canine''. 

\begin{figure}
	\centering
	\begin{subfigure}{0.48\linewidth}
		\includegraphics[width=1.0\columnwidth]{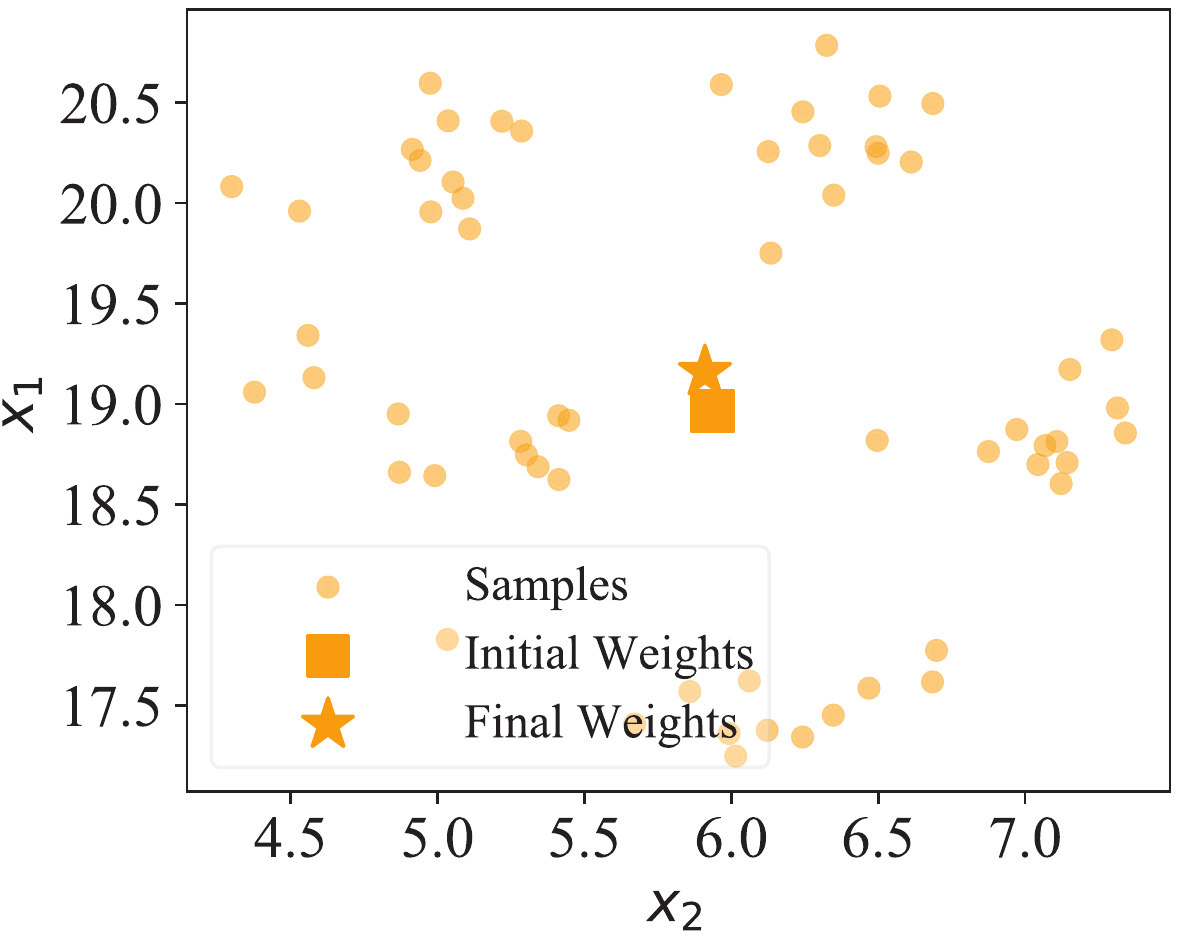}
		\caption{The root nodes.}
		\label{fig7a}
	\end{subfigure}
	\hfill
	\begin{subfigure}{0.48\linewidth}
		\includegraphics[width=1.0\columnwidth]{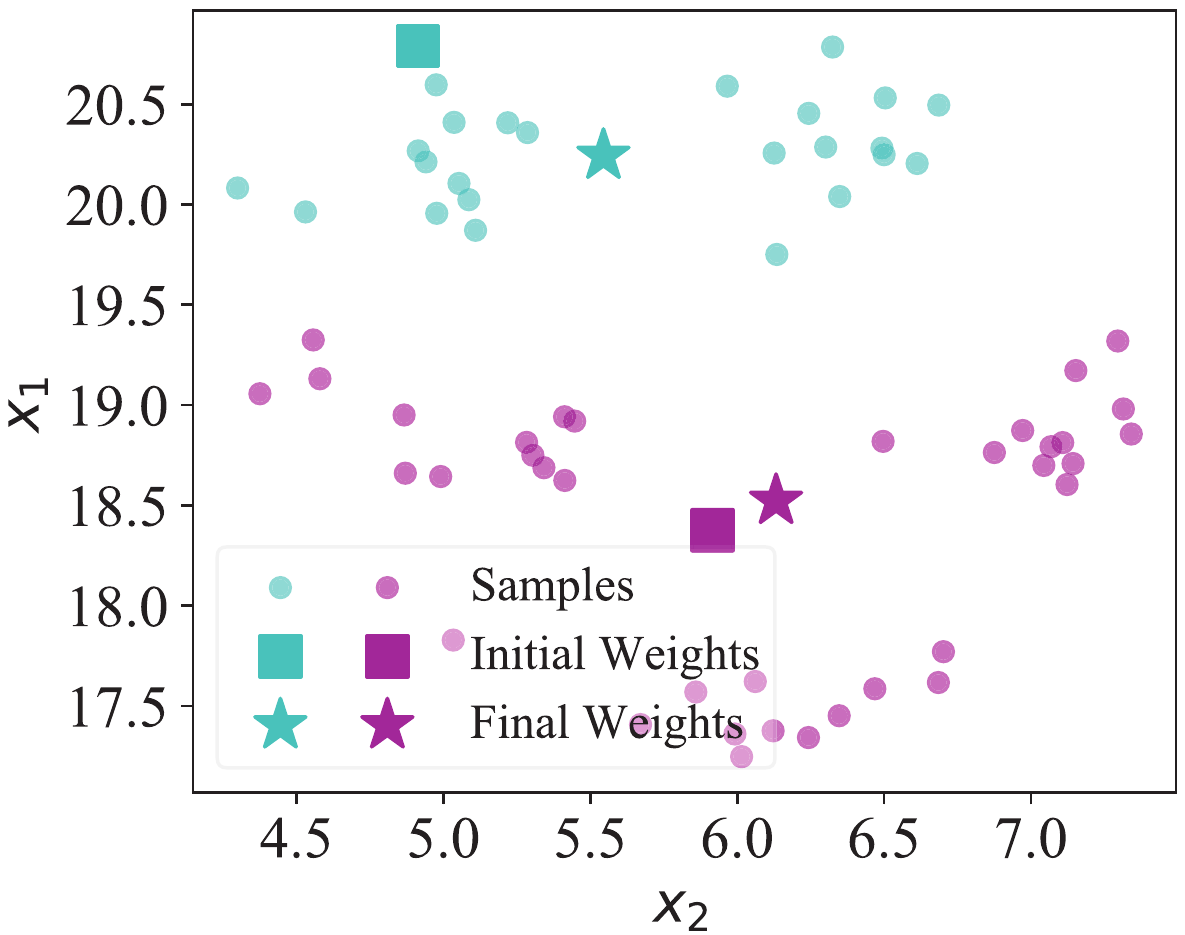}
		\caption{The intermediate nodes.}
		\label{fig7b}
	\end{subfigure}
	\hfill
	\begin{subfigure}{0.48\linewidth}
		\includegraphics[width=1.0\columnwidth]{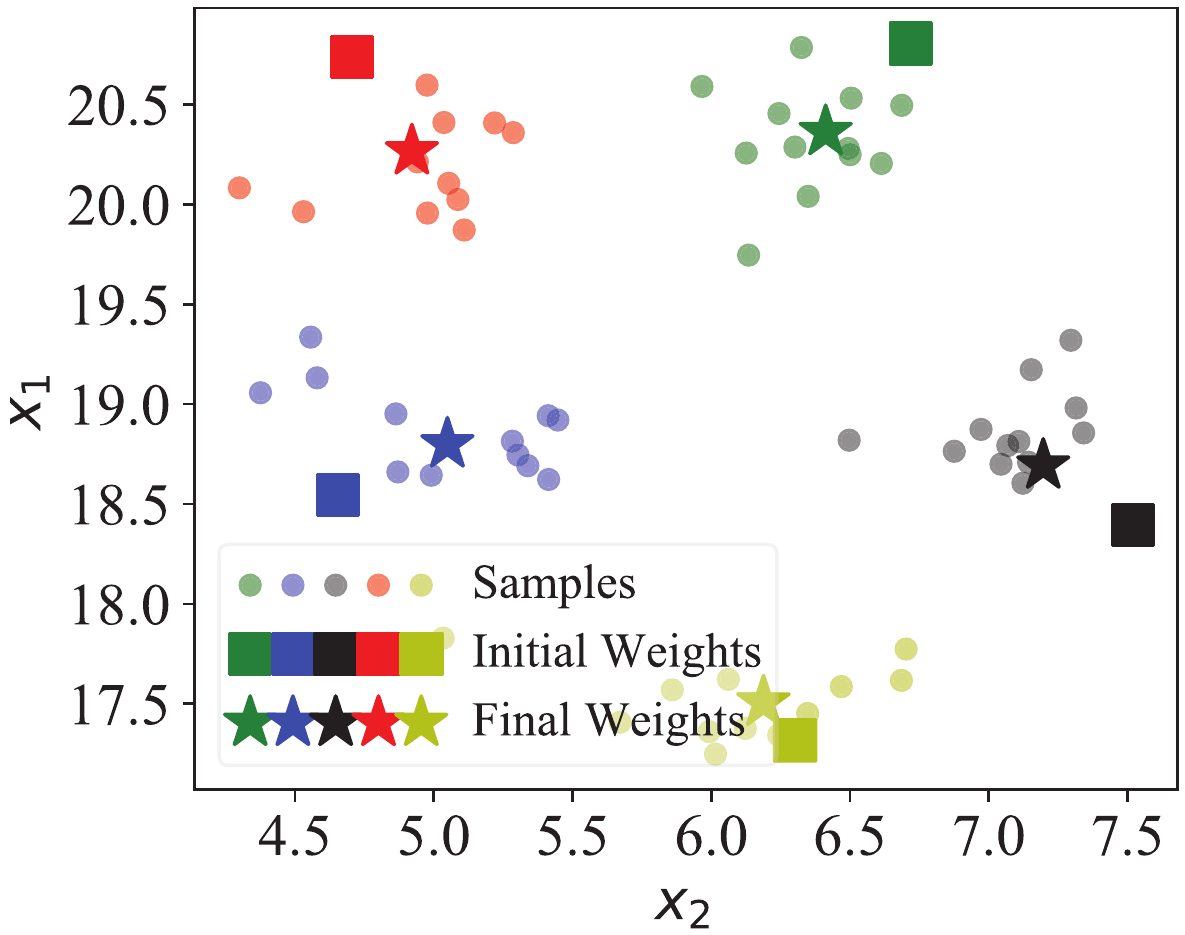}
		\caption{The leaf nodes.}
		\label{fig7c}
	\end{subfigure}
	\hfill
	\begin{subfigure}{0.48\linewidth}
		\includegraphics[width=1.0\columnwidth]{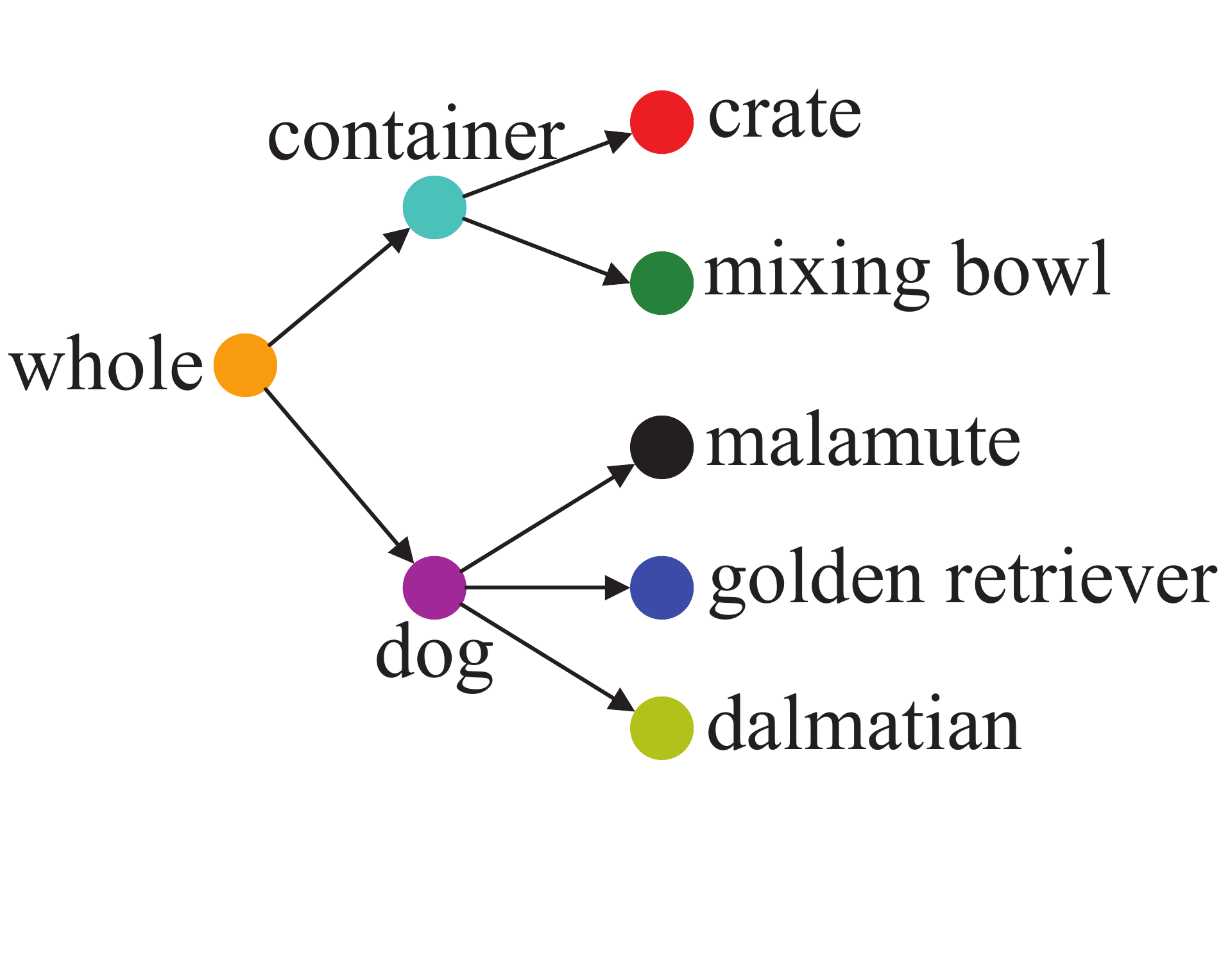}
		\caption{Structure of decision trees.}
		\label{fig7d}
	\end{subfigure}
	\caption{Visualization of the weights of decison trees.}
	\label{fig7}
\end{figure}

\noindent \textbf{How does our MetaDT work?} In Figures~\ref{fig7a}, \ref{fig7b}, and \ref{fig7c}, we visualize the weights of decision tree on a 5-way 1-shot task of miniImagenet. Here, the weights of decision tree before and after fast adaptation are marked by the squares and stars, respectively. Besides, the structure of decison tree is shown in Figure~\ref{fig7d}. We find that the final weights (\emph{i.e.}, final prototypes) marked by stars are much closer to the centers of few-shot classes/superclass after fast adaptation. This visualization indicates that our MetaDT effectively learns the class hierarchy and infers the task-specific decision tree.

\section{Conclusion}

In this paper, we propose a novel decision tree-based meta-learning framework for interpretable few-shot learning, called MetaDT. In particular, we replace the black-box FSL classifier with an interpretable decision tree. In addition, a novel decision tree inference network and a two-loop optimization mechanism is designed, respectively, for a fast adaptation of decision tree. Extensive experiments and interpretability analyses show our MetaDT is effective and provides interpretability for novel class predictions.

{\small
\bibliographystyle{ieee_fullname}
\bibliography{egbib}

\begin{thebibliography}{10}\itemsep=-1pt

\bibitem{altae2017low}
Han Altae-Tran, Bharath Ramsundar, Aneesh~S Pappu, and Vijay Pande.
\newblock Low data drug discovery with one-shot learning.
\newblock {\em ACS central science}, 3(4):283--293, 2017.

\bibitem{baik2021meta}
Sungyong Baik, Janghoon Choi, Heewon Kim, Dohee Cho, Jaesik Min, and Kyoung~Mu
  Lee.
\newblock Meta-learning with task-adaptive loss function for few-shot learning.
\newblock In {\em ICCV}, pages 9465--9474, 2021.

\bibitem{BaikCCKL20}
Sungyong Baik, Myungsub Choi, Janghoon Choi, Heewon Kim, and Kyoung~Mu Lee.
\newblock Meta-learning with adaptive hyperparameters.
\newblock In {\em NeurIPS}, 2020.

\bibitem{banegas2021towards}
Antonio~Jes{\'u}s Banegas-Luna, Jorge Pe{\~n}a-Garc{\'\i}a, Adrian Iftene,
  Fiorella Guadagni, Patrizia Ferroni, Noemi Scarpato, Fabio~Massimo Zanzotto,
  Andr{\'e}s Bueno-Crespo, and Horacio P{\'e}rez-S{\'a}nchez.
\newblock Towards the interpretability of machine learning predictions for
  medical applications targeting personalised therapies: A cancer case survey.
\newblock {\em International Journal of Molecular Sciences}, 22(9):4394, 2021.

\bibitem{BertinettoHTV19}
Luca Bertinetto, Jo{\~{a}}o~F. Henriques, Philip H.~S. Torr, and Andrea
  Vedaldi.
\newblock Meta-learning with differentiable closed-form solvers.
\newblock In {\em ICLR}, 2019.

\bibitem{CaoBL21}
Kaidi Cao, Maria Brbic, and Jure Leskovec.
\newblock Concept learners for few-shot learning.
\newblock In {\em ICLR}, 2021.

\bibitem{carvalho2019machine}
Diogo~V Carvalho, Eduardo~M Pereira, and Jaime~S Cardoso.
\newblock Machine learning interpretability: A survey on methods and metrics.
\newblock {\em Electronics}, 8(8):832, 2019.

\bibitem{ChenLKWH19}
Wei{-}Yu Chen, Yen{-}Cheng Liu, Zsolt Kira, Yu{-}Chiang~Frank Wang, and
  Jia{-}Bin Huang.
\newblock A closer look at few-shot classification.
\newblock In {\em ICLR}, 2019.

\bibitem{chen2020new}
Yinbo Chen, Xiaolong Wang, Zhuang Liu, Huijuan Xu, Trevor Darrell, et~al.
\newblock A new meta-baseline for few-shot learning.
\newblock In {\em ICML}, 2020.

\bibitem{ChenFZJXS19}
Zitian Chen, Yanwei Fu, Yinda Zhang, Yu{-}Gang Jiang, Xiangyang Xue, and Leonid
  Sigal.
\newblock Multi-level semantic feature augmentation for one-shot learning.
\newblock {\em {IEEE} Trans. Image Process.}, 28(9):4594--4605, 2019.

\bibitem{Chen_2021_CVPR}
Zhengyu Chen, Jixie Ge, Heshen Zhan, Siteng Huang, and Donglin Wang.
\newblock Pareto self-supervised training for few-shot learning.
\newblock In {\em CVPR}, pages 13663--13672, 2021.

\bibitem{elhoseiny2019creativity}
Mohamed Elhoseiny and Mohamed Elfeki.
\newblock Creativity inspired zero-shot learning.
\newblock In {\em ICCV}, pages 5784--5793, 2019.

\bibitem{finn2017model}
Chelsea Finn, Pieter Abbeel, Sergey Levine, et~al.
\newblock Model-agnostic meta-learning for fast adaptation of deep networks.
\newblock In {\em ICML}, pages 1126--1135, 2017.

\bibitem{FromeCSBDRM13}
Andrea Frome, Gregory~S. Corrado, Jonathon Shlens, Samy Bengio, Jeffrey Dean,
  Marc'Aurelio Ranzato, and Tomas Mikolov.
\newblock Devise: {A} deep visual-semantic embedding model.
\newblock In {\em NeurIPS}, pages 2121--2129, 2013.

\bibitem{gao2021curvature}
Zhi Gao, Yuwei Wu, Yunde Jia, and Mehrtash Harandi.
\newblock Curvature generation in curved spaces for few-shot learning.
\newblock In {\em ICCV}, pages 8691--8700, 2021.

\bibitem{he2016deep}
Kaiming He, Xiangyu Zhang, Shaoqing Ren, and Jian Sun.
\newblock Deep residual learning for image recognition.
\newblock In {\em CVPR}, pages 770--778, 2016.

\bibitem{kampffmeyer2019rethinking}
Michael Kampffmeyer, Yinbo Chen, Xiaodan Liang, Hao Wang, Yujia Zhang, and
  Eric~P Xing.
\newblock Rethinking knowledge graph propagation for zero-shot learning.
\newblock In {\em CVPR}, pages 11487--11496, 2019.

\bibitem{kontschieder2015deep}
Peter Kontschieder, Madalina Fiterau, Antonio Criminisi, and Samuel~Rota Bulo.
\newblock Deep neural decision forests.
\newblock In {\em ICCV}, pages 1467--1475, 2015.

\bibitem{KontschiederFCB16}
Peter Kontschieder, Madalina Fiterau, Antonio Criminisi, and Samuel~Rota
  Bul{\`{o}}.
\newblock Deep neural decision forests.
\newblock In {\em IJCAI}, pages 4190--4194, 2016.

\bibitem{lavanya2012ensemble}
D Lavanya and K~Usha Rani.
\newblock Ensemble decision tree classifier for breast cancer data.
\newblock {\em International Journal of Information Technology Convergence and
  Services}, 2(1):17--24, 2012.

\bibitem{lee2019meta}
Kwonjoon Lee, Subhransu Maji, Avinash Ravichandran, and Stefano Soatto.
\newblock Meta-learning with differentiable convex optimization.
\newblock In {\em CVPR}, pages 10657--10665, 2019.

\bibitem{boostingfew}
Aoxue Li, Weiran Huang, Xu Lan, Jiashi Feng, Zhenguo Li, and Liwei Wang.
\newblock Boosting few-shot learning with adaptive margin loss.
\newblock In {\em CVPR}, pages 12576--12584, 2020.

\bibitem{LiXHWGL19}
Wenbin Li, Jinglin Xu, Jing Huo, Lei Wang, Yang Gao, and Jiebo Luo.
\newblock Distribution consistency based covariance metric networks for
  few-shot learning.
\newblock In {\em AAAI}, pages 8642--8649, 2019.

\bibitem{liu2020negative}
Bin Liu, Yue Cao, Yutong Lin, Qi Li, Zheng Zhang, Mingsheng Long, and Han Hu.
\newblock Negative margin matters: Understanding margin in few-shot
  classification.
\newblock In {\em ECCV}, pages 438--455, 2020.

\bibitem{Mangla0SKBK20}
Puneet Mangla, Mayank Singh, Abhishek Sinha, Nupur Kumari, Vineeth~N.
  Balasubramanian, and Balaji Krishnamurthy.
\newblock Charting the right manifold: Manifold mixup for few-shot learning.
\newblock In {\em WACV}, pages 2207--2216, 2020.

\bibitem{miller1998wordnet}
George~A Miller.
\newblock {\em WordNet: An electronic lexical database}.
\newblock MIT press, 1998.

\bibitem{zhimao2019few}
Zhimao Peng, Zechao Li, Junge Zhang, Yan Li, Guo{-}Jun Qi, and Jinhui Tang.
\newblock Few-shot image recognition with knowledge transfer.
\newblock In {\em ICCV}, pages 441--449, 2019.

\bibitem{pennington2014glove}
Jeffrey Pennington, Richard Socher, and Christopher Manning.
\newblock Glove: Global vectors for word representation.
\newblock In {\em EMNLP}, pages 1532--1543, 2014.

\bibitem{RenTRSSTLZ18}
Mengye Ren, Eleni Triantafillou, Sachin Ravi, Jake Snell, Kevin Swersky,
  Joshua~B. Tenenbaum, Hugo Larochelle, and Richard~S. Zemel.
\newblock Meta-learning for semi-supervised few-shot classification.
\newblock In {\em ICLR}, 2019.

\bibitem{Rizve_2021_CVPR}
Mamshad~Nayeem Rizve, Salman Khan, Fahad~Shahbaz Khan, and Mubarak Shah.
\newblock Exploring complementary strengths of invariant and equivariant
  representations for few-shot learning.
\newblock In {\em CVPR}, pages 10836--10846, 2021.

\bibitem{rodriguez2020embedding}
Pau Rodr{\'\i}guez, Issam Laradji, Alexandre Drouin, and Alexandre Lacoste.
\newblock Embedding propagation: Smoother manifold for few-shot classification.
\newblock In {\em ECCV}, 2020.

\bibitem{rusu2018meta}
Andrei~A Rusu, Dushyant Rao, Jakub Sygnowski, Oriol Vinyals, Razvan Pascanu,
  Simon Osindero, and Raia Hadsell.
\newblock Meta-learning with latent embedding optimization.
\newblock In {\em ICLR}, 2018.

\bibitem{safavian1991survey}
S~Rasoul Safavian and David Landgrebe.
\newblock A survey of decision tree classifier methodology.
\newblock {\em IEEE transactions on systems, man, and cybernetics},
  21(3):660--674, 1991.

\bibitem{satorras2018few}
Victor~Garcia Satorras and Joan~Bruna Estrach.
\newblock Few-shot learning with graph neural networks.
\newblock In {\em ICLR}, 2018.

\bibitem{babysteps2019}
Eli Schwartz, Leonid Karlinsky, Rog{\'{e}}rio~Schmidt Feris, Raja Giryes, and
  Alexander~M. Bronstein.
\newblock Baby steps towards few-shot learning with multiple semantics.
\newblock {\em CoRR}, abs/1906.01905, 2019.

\bibitem{snell2017prototypical}
Jake Snell, Kevin Swersky, Richard Zemel, et~al.
\newblock Prototypical networks for few-shot learning.
\newblock In {\em NeurIPS}, pages 4077--4087, 2017.

\bibitem{Tang_2021_CVPR}
Shixiang Tang, Dapeng Chen, Lei Bai, Kaijian Liu, Yixiao Ge, and Wanli Ouyang.
\newblock Mutual crf-gnn for few-shot learning.
\newblock In {\em CVPR}, pages 2329--2339, 2021.

\bibitem{tian2020rethinking}
Yonglong Tian, Yue Wang, Dilip Krishnan, Joshua~B Tenenbaum, and Phillip Isola.
\newblock Rethinking few-shot image classification: a good embedding is all you
  need?
\newblock In {\em ECCV}, pages 266--282, 2020.

\bibitem{VartakTMBL17}
Manasi Vartak, Arvind Thiagarajan, Conrado Miranda, Jeshua Bratman, and Hugo
  Larochelle.
\newblock A meta-learning perspective on cold-start recommendations for items.
\newblock In {\em NeurIPS}, pages 6904--6914, 2017.

\bibitem{vinyals2016matching}
Oriol Vinyals, Charles Blundell, Timothy Lillicrap, Daan Wierstra, et~al.
\newblock Matching networks for one shot learning.
\newblock In {\em NeurIPS}, pages 3630--3638, 2016.

\bibitem{AlvinWan21}
Alvin Wan, Lisa Dunlap, Daniel Ho, Jihan Yin, Scott Lee, Henry Jin, Suzanne
  Petryk, Sarah~Adel Bargal, and Joseph~E. Gonzalez.
\newblock Nbdt: Neural-backed decision trees.
\newblock In {\em ICLR}, 2021.

\bibitem{WanCLYZY019}
Ziyu Wan, Dongdong Chen, Yan Li, Xingguang Yan, Junge Zhang, Yizhou Yu, and
  Jing Liao.
\newblock Transductive zero-shot learning with visual structure constraint.
\newblock In {\em NeurIPS}, pages 9972--9982, 2019.

\bibitem{WangLVNKN21}
Bowen Wang, Liangzhi Li, Manisha Verma, Yuta Nakashima, Ryo Kawasaki, and
  Hajime Nagahara.
\newblock Mtunet: Few-shot image classification with visual explanations.
\newblock In {\em CVPR Workshops}, pages 2294--2298, 2021.

\bibitem{WangYKN20}
Yaqing Wang, Quanming Yao, James~T. Kwok, and Lionel~M. Ni.
\newblock Generalizing from a few examples: {A} survey on few-shot learning.
\newblock {\em {ACM} Comput. Surv.}, 53(3):63:1--63:34, 2020.

\bibitem{Wertheimer_2021_CVPR}
Davis Wertheimer, Luming Tang, and Bharath Hariharan.
\newblock Few-shot classification with feature map reconstruction networks.
\newblock In {\em CVPR}, pages 8012--8021, 2021.

\bibitem{xing2019adaptive}
Chen Xing, Negar Rostamzadeh, Boris~N Oreshkin, and Pedro~O Pinheiro.
\newblock Adaptive cross-modal few-shot learning.
\newblock In {\em NeurIPS}, pages 4848--4858, 2019.

\bibitem{0001FLWLH0X21}
Chengming Xu, Yanwei Fu, Chen Liu, Chengjie Wang, Jilin Li, Feiyue Huang, Li
  Zhang, and Xiangyang Xue.
\newblock Learning dynamic alignment via meta-filter for few-shot learning.
\newblock In {\em CVPR}, pages 5182--5191, 2021.

\bibitem{xue2020region}
Zhiyu Xue, Lixin Duan, Wen Li, Lin Chen, and Jiebo Luo.
\newblock Region comparison network for interpretable few-shot image
  classification.
\newblock {\em arXiv preprint arXiv:2009.03558}, 2020.

\bibitem{YangSG19}
Bin{-}Bin Yang, Song{-}Qing Shen, and Wei Gao.
\newblock Weighted oblique decision trees.
\newblock In {\em AAAI}, pages 5621--5627, 2019.

\bibitem{distributionpropagation2020}
Ling Yang, Liangliang Li, Zilun Zhang, Xinyu Zhou, Erjin Zhou, and Yu Liu.
\newblock {DPGN:} distribution propagation graph network for few-shot learning.
\newblock In {\em CVPR}, pages 13390--13399, 2020.

\bibitem{yang2018deep}
Yongxin Yang, Irene~Garcia Morillo, and Timothy~M Hospedales.
\newblock Deep neural decision trees.
\newblock {\em arXiv preprint arXiv:1806.06988}, 2018.

\bibitem{YeHZS20}
Han{-}Jia Ye, Hexiang Hu, De{-}Chuan Zhan, and Fei Sha.
\newblock Few-shot learning via embedding adaptation with set-to-set functions.
\newblock In {\em CVPR}, pages 8805--8814, 2020.

\bibitem{zhang2021metanode}
Baoquan Zhang, Xutao Li, Yunming Ye, Shanshan Feng, and Rui Ye.
\newblock Metanode: Prototype optimization as a neural ode for few-shot
  learning.
\newblock {\em arXiv preprint arXiv:2103.14341}, 2021.

\bibitem{zhang2021prototype}
Baoquan Zhang, Xutao Li, Yunming Ye, Zhichao Huang, and Lisai Zhang.
\newblock Prototype completion with primitive knowledge for few-shot learning.
\newblock In {\em CVPR}, pages 3754--3762, 2021.

\bibitem{zhang2020deepemd}
Chi Zhang, Yujun Cai, Guosheng Lin, and Chunhua Shen.
\newblock Deepemd: Few-shot image classification with differentiable earth
  mover's distance and structured classifiers.
\newblock In {\em CVPR}, pages 12203--12213, 2020.

\bibitem{ZhangZNXY19}
Jian Zhang, Chenglong Zhao, Bingbing Ni, Minghao Xu, and Xiaokang Yang.
\newblock Variational few-shot learning.
\newblock In {\em ICCV}, pages 1685--1694, 2019.

\end{thebibliography}
}

\end{document}